\documentclass{article}
\usepackage[preprint]{colm2026_conference}
\usepackage{microtype}
\usepackage{hyperref}
\usepackage{url}
\usepackage{booktabs}
\usepackage{graphicx}
\usepackage{amsmath}
\usepackage{amssymb}
\usepackage{subcaption}
\usepackage{multirow}
\usepackage{lineno}
\definecolor{darkblue}{rgb}{0, 0, 0.5}
\hypersetup{colorlinks=true, citecolor=darkblue, linkcolor=darkblue, urlcolor=darkblue}

\title{Capability-Routed Visual Retrieval and Evidence Threading for Long-Context Document Question Answering}

\author{
Amirul Rahman, Aisha Karim, Kenji Nakamura, Yi-Fan Ng \\
University of Malaya \\
\texttt{kenjinakamura@um.edu.my}
}

\begin{document}
\ifcolmsubmission
\linenumbers
\fi
\maketitle

\begin{abstract}
Annual reports, diligence packs, and infographic dashboards bury numbers in page images: axes, cell grids, and footnotes that OCR pipelines flatten and that page-level visual retrievers still treat as interchangeable in-context examples. We keep a frozen Qwen2.5-VL-7B-Instruct generator and a ColPali / VisRAG-Ret page index, and insert three modules. A capability-aware visual router (CAVR) tags each retrieved page as text, table, chart, layout, or mixed and mixes specialist experts before generation. Weak-to-strong page selection (WSPS) distils a frozen 7B answerability teacher into a 3B selection head so ranking is no longer a single InfoNCE score. Visual evidence threading (VET) builds layout-anchored paths of length at most three and lets the generator read the thread rather than a flat top-$k$ list. On gold-page DocVQA / ChartQA / InfographicVQA the 7B system reaches 96.3 / 90.1 / 85.4. Under the VisRAG top-3 protocol the mean generation accuracy is 62.74 versus 59.39 for the same backbone with concatenation. On MMLongBench-Doc retrieve-then-read, F1 moves from 19.2 to 22.6 and multi-page accuracy from 16.4 to 21.2. ViDoRe nDCG@5 after WSPS reranking is 83.6, with TAT-DQA financial reports at 70.4.
\end{abstract}

\section{Introduction}

Financial filings and business PDFs are not natural photographs. A query about operating margin may need a line chart on one page, a consolidated table two pages later, and a polarity-reversing footnote in the notes. Text retrieval-augmented generation (RAG) first runs optical character recognition (OCR) and then embeds chunks~\cite{lewis2020rag,karpukhin2020dpr,robertson2009bm25}. Layout, axis ticks, and cell alignment die in that parse. Vision-first retrievers such as ColPali and VisRAG keep the page as an image~\cite{faysse2025colpali,yu2025visrag}, which is the right unit for annual reports, yet the generator still concatenates the top-$k$ pages as if they were homogeneous visual prompts.

\cite{zhou2024visual} showed that a handful of visual in-context examples can steer a large vision--language model (VLM). That result assumes the examples share a skill and that order is secondary. Adjacent pages in a 10-K do not: one is a bar chart, one is a dense legal opinion, one is a split balance sheet. Concatenating them is closer to stuffing a chaotic context than to visual in-context learning. On VisRAG, MiniCPM-V 2.0 ChartQA accuracy falls from 34.92 at top-1 to 22.22 at top-3 page concatenation~\cite{yu2025visrag,yao2024minicpmv}, so ``more pages as more shots'' is already false. MMLongBench-Doc further shows that even GPT-4o scores only 44.9 F1 when it reads whole PDFs of 47.5 pages on average, and open LVLMs that concatenate every page often trail OCR-then-LLM stacks~\cite{ma2024mmlongbench,openai2024gpt4o}.

Three structural gaps remain after 2025 visual document RAG. Retrievers score semantic near-neighbours; they do not say whether a page should be read by an OCR expert, a table merger, or a chart-axis head~\cite{hu2024docowl15,masry-etal-2022-chartqa}. A 3B late-interaction index and a 7B reader are not the same model, so contrastive ranks can hide the page that the reader could actually use. Cross-page evidence is a visual stream of running headers, cropped tables, and captions that live on a different folio than the figure---a setting that layout-aware graphs and agentic planners attack with heavier controllers~\cite{sourati2026ladrag,magerag2026,mmr22026}, not with a short, layout-anchored thread that a frozen 7B decoder can consume.

\begin{figure}[!t]
\centering
\includegraphics[width=\linewidth]{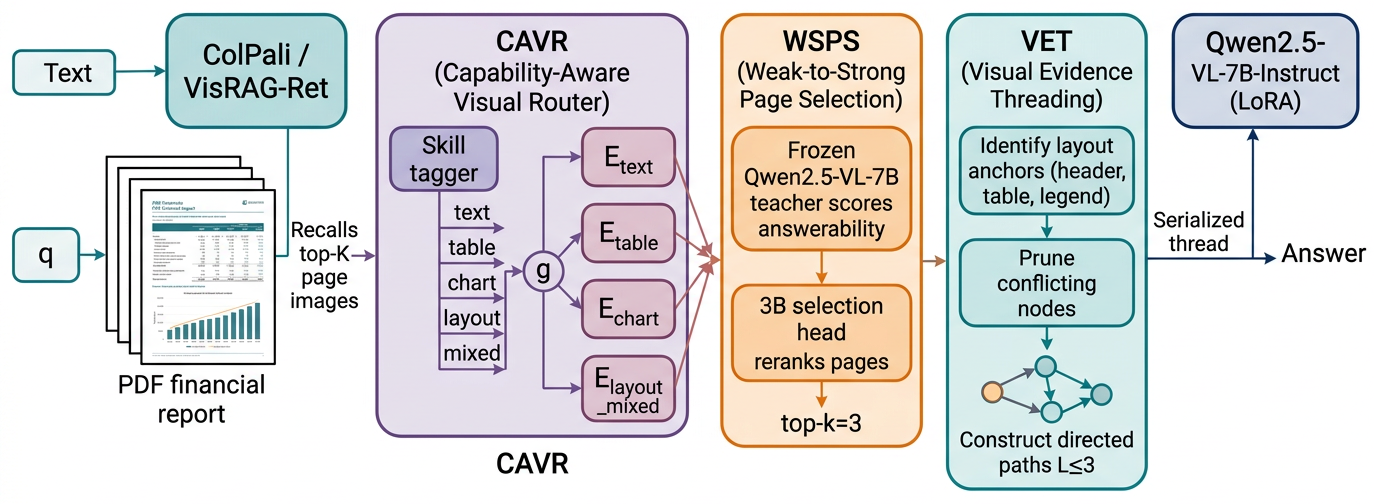}
\caption{Ours (CAVR+WSPS+VET). ColPali / VisRAG-Ret recall page images. CAVR routes each page through text, table, chart, and layout experts. WSPS reranks with a frozen 7B teacher. VET builds layout-anchored threads that Qwen2.5-VL-7B-Instruct reads under LoRA.}
\label{fig:arch}
\end{figure}

We freeze Qwen2.5-VL-7B-Instruct~\cite{bai2025qwen25vl} and a ColPali or VisRAG-Ret index, and insert three modules (Figure~\ref{fig:arch} and Section~\ref{sec:method}): capability-aware visual routing (CAVR), weak-to-strong page selection (WSPS), and visual evidence threading (VET). Contributions:
\begin{itemize}
\item CAVR classifies each retrieved page into text, table, chart, layout, or mixed skill and mixes four light experts so heterogeneous pages are no longer one visual channel.
\item WSPS trains a selection head on frozen 7B answerability labels, then interpolates that head with late-interaction retrieval.
\item VET compiles layout anchors into directed paths of length $L{\le}3$ and serialises the winning path. Ours (CAVR+WSPS+VET) reaches 96.3 / 90.1 / 85.4 on gold-page DocVQA / ChartQA / InfoVQA, 62.74 mean VisRAG-protocol accuracy, 22.6 MMLongBench-Doc F1, and 83.6 ViDoRe nDCG@5.
\end{itemize}

\section{Related Work}

\subsection{Document VLMs and visual in-context learning}
OCR-free readers---Donut, Pix2Struct, UReader, Monkey, CogAgent, DocOwl 1.5---treat a page as a screenshot and emit answers or structured markup~\cite{kim2022donut,lee2023pix2struct,ye2023ureader,liu2023monkey,hong2024cogagent,hu2024docowl15}. General VLMs (Qwen-VL, Qwen2-VL, Qwen2.5-VL, InternVL, LLaVA-OneVision, PaliGemma, MiniCPM-V) raise DocVQA into the mid-90s at 7B--72B~\cite{bai2023qwenvl,wang2024qwen2vl,bai2025qwen25vl,chen2024internvl,chen2024internvl25,li2024llavaov,beyer2024paligemma,yao2024minicpmv}. Chart pretraining such as MatCha derenders plots into math~\cite{liu-etal-2023-matcha}; knowledge-augmented VL transformers inject text facts beside pixels~\cite{gui-etal-2022-kat}. Modular VQA compiles a question into a program over specialists~\cite{subramanian-etal-2023-modular}, which is the right \emph{shape} for heterogeneous pages, but those programs assume cropped natural images, not retrieved PDF folios. Visual ICL concatenates labelled images; \cite{zhou2025weak} show that weak-to-strong generalisation fails when one teacher signal mixes many capabilities. We take that multi-capability diagnosis into visual RAG: chart reading, table alignment, and layout localisation are not one skill, so they should not share one projection.

\subsection{Visual document RAG and late interaction}
Text RAG for knowledge-intensive QA is mature~\cite{lewis2020rag,zhang-etal-2023-double,saad-falcon-etal-2024-ares}. Visual RAG attaches an image to the query~\cite{lin-byrne-2022-retrieval,ramos-etal-2023-retrieval,chen-etal-2023-pre-trained}; medical rationale-guided RAG insists that answers cite retrieved evidence~\cite{sohn-etal-2025-rationale}. Multilingual RAG in 2026 is still largely textual~\cite{ranaldi-etal-2026-multilingual}. On documents, ColBERT-style late interaction~\cite{khattab2020colbert} moved to page patches in ColPali~\cite{faysse2025colpali,beyer2024paligemma} and to whole-page VisRAG-Ret~\cite{yu2025visrag}. SigLIP, BiPali, and OCR+BGE pipelines are the standard ViDoRe baselines~\cite{zhai2023siglip,xiao2023cpack,lee2024nvembed}. 2026 systems scale the same late-interaction idea: Nemotron ColEmbed V2 tops ViDoRe V3~\cite{nemotron2026}; AnchorFold and ColSNAP compress patch indexes~\cite{anchorfold2026,colsnap2026}; PULSAR pools ColPali-style vectors for enterprise decks~\cite{pulsar2026}. MARA reweights regions by the query~\cite{mara2026}; LAD-RAG adds a symbolic layout graph and an agent~\cite{sourati2026ladrag}; MAGE-RAG builds a page--element graph and a budgeted controller~\cite{magerag2026}; MM-R2 plans \emph{what} and \emph{where} before search~\cite{mmr22026}. Those stacks either enlarge the retriever or wrap an LLM agent around it. We keep the 3B index and change how pages are routed, selected, and threaded for a 7B reader.

Long-context alignment further shows that not every token in a long window is equally useful: GATEAU selects influential samples before alignment~\cite{si2025gateau}. Page selection is the visual analogue---a retrieved folio can be topically close and still unusable for answering.

\subsection{Long PDFs, threads, and faithfulness}
DocVQA, ChartQA, InfographicVQA, and MMLongBench-Doc define the evaluation surface we use~\cite{mathew2021docvqa,masry-etal-2022-chartqa,mathew2022infovqa,ma2024mmlongbench}. Cross-page items are 33.7\% of MMLongBench-Doc; financial reports are 8.1\% of its documents. Concatenating every page, or packing five pages into one canvas, cuts the thread. We want a short directed path with layout anchors, not a free-form agent trace and not a second copy of visual ICL. CLIP-style dual encoders remain a useful single-vector reference~\cite{radford2021clip} but lose the patch matches ColPali keeps.

\section{Method}
\label{sec:method}

A text query $q$ and a corpus $\mathcal{P}$ of page images map to an answer $y$. Figure~\ref{fig:arch} shows the flow. Retrieval is frozen ColPali or VisRAG-Ret; generation is Qwen2.5-VL-7B-Instruct with LoRA~\cite{hu2022lora,bai2025qwen25vl}. CAVR, WSPS, and VET are the only trained modules.

\subsection{Problem formulation}
Let $\mathcal{D}_K\subset\mathcal{P}$ be the top-$K$ pages under late interaction ($K{=}20$). CAVR emits page representations $\{h_i\}$, WSPS keeps a subset $\mathcal{D}_k$ with $k{=}3$, and VET returns a thread $T$. Decoding is
\begin{align}
p(y\mid q,\mathcal{P})=\mathrm{Dec}_{\phi}\bigl(q,\,T(q,\mathcal{D}_k)\bigr).
\label{eq:task}
\end{align}
The decoder in~\eqref{eq:task} never sees the raw top-$k$ list. Oracle pages, when used in analysis, replace $\mathcal{D}_k$ but still pass CAVR and VET.

\subsection{Retriever}
ColPali encodes $q$ as token vectors $\{q_t\}$ and a page $P$ as patch vectors $\{p_j\}$. Late-interaction MaxSim~\cite{khattab2020colbert,faysse2025colpali} is
\begin{align}
s_{\mathrm{ret}}(q,P)=\sum_{t}\max_{j}\langle q_t,p_j\rangle.
\label{eq:maxsim}
\end{align}
VisRAG-Ret substitutes a pooled page vector for $\{p_j\}$ on the VisRAG protocol~\cite{yu2025visrag}. We do not fine-tune~\eqref{eq:maxsim}; WSPS only reranks.

\subsection{CAVR: capability-aware visual router}
Visual ICL assumes homogeneous shots. Document pages do not. A five-way linear head on the ColPali patch mean predicts a skill
\begin{align}
r_i=\mathrm{softmax}(W_r\,\bar{p}_i+b_r)\in\Delta^{4},
\label{eq:skill}
\end{align}
with classes $\{\mathrm{text},\mathrm{table},\mathrm{chart},\mathrm{layout},\mathrm{mixed}\}$ in~\eqref{eq:skill}. Supervision comes from dataset metadata (ChartQA$\to$chart, TAT-DQA / WikiTableQuestions$\to$table, dense DocVQA pages$\to$text, InfographicVQA$\to$mixed) and DocOwl-style structure tags~\cite{hu2024docowl15}. Four experts then read $P_i$: $E_{\mathrm{text}}$ is the Qwen2.5-VL dynamic-resolution ViT; $E_{\mathrm{table}}$ merges horizontal neighbour patches; $E_{\mathrm{chart}}$ concatenates axis-tick OCR; $E_{\mathrm{layout}}$ keeps region bounding tokens. A query-conditioned gate
\begin{align}
g_i=\mathrm{softmax}\bigl(W_g[r_i;q]\bigr)
\label{eq:gate}
\end{align}
mixes experts in~\eqref{eq:hpage}:
\begin{align}
h_i=\sum_{e}g_{i,e}\,E_e(P_i).
\label{eq:hpage}
\end{align}
The generator prompt prefixes the argmax skill from~\eqref{eq:gate}. Ablating CAVR forces every page through $E_{\mathrm{text}}$. Mixing experts is cheaper than a sparse mixture-of-experts language model~\cite{jiang2024mixtral}: the gate is a $5{\times}4$ matrix on a 7B reader, not a 8$\times$22B decoder.

\subsection{WSPS: weak-to-strong page selection}
A single InfoNCE rank cannot lift chart reading and clause localisation together~\cite{zhou2025weak}. We freeze the 7B generator and, for each candidate $P_i$ in $\mathcal{D}_K$, record whether the gold answer is extractable from $(q,P_i)$ under relaxed match (ChartQA-style $\pm 5\%$ on numbers~\cite{masry-etal-2022-chartqa}). Gold evidence pages are positives. The teacher probability and hard label are
\begin{align}
\pi_i=p_{\phi}(\mathrm{span}(y)\mid q,P_i),\qquad
y_i=\mathbf{1}[\pi_i\ge\tau]\lor\mathbf{1}[P_i\in\mathcal{P}^\star],
\label{eq:teacher}
\end{align}
with $\tau{=}0.35$ in~\eqref{eq:teacher}. A linear head $s_{\mathrm{sel}}(q,P_i)$ on the retriever's late-interaction summary is trained with
\begin{align}
\mathcal{L}_{\mathrm{WSPS}}=\mathrm{BCE}(s_{\mathrm{sel}},y)+\beta\,\mathrm{KL}(s_{\mathrm{sel}}\Vert\pi),\qquad \beta=0.3.
\label{eq:lwsps}
\end{align}
The 7B weights are not updated in~\eqref{eq:lwsps}. Test-time fusion is
\begin{align}
s(q,P_i)=\lambda_{\mathrm{ret}}s_{\mathrm{ret}}+\lambda_{\mathrm{sel}}s_{\mathrm{sel}}(q,P_i),
\label{eq:fuse}
\end{align}
with default $\lambda_{\mathrm{sel}}{=}0.4$ in~\eqref{eq:fuse} and $\lambda_{\mathrm{ret}}{=}1-\lambda_{\mathrm{sel}}$. Per-skill temperatures $\tau_e$ on $s_{\mathrm{sel}}$ stop dense-text pages from crowding out charts:
\begin{align}
\tilde{s}_{\mathrm{sel}}=\tau_{e(i)}^{-1}s_{\mathrm{sel}}(q,P_i).
\label{eq:temp}
\end{align}
\eqref{eq:temp} is a calibration, not a second retriever. Contextual faithfulness training that rewrites the generator~\cite{DBLP:conf/aaai/SiZGBWGLLHCQZCS26} is complementary: we change which pages enter the window, not the RL objective.

\subsection{VET: visual evidence threading}
\cite{zhou2023thread} insert a ``first summarise, then reason'' step in chaotic \emph{text}. Multi-page PDFs are chaotic because headers repeat, tables split across folios, and captions drift from figures. We build a directed graph whose nodes are pages plus layout anchors (title block, header row, legend, page number) as in~\eqref{eq:node},
\begin{align}
v_i=(P_i,a_i,r_i),\qquad a_i=\mathrm{Anchor}(P_i).
\label{eq:node}
\end{align}
Edge weights in~\eqref{eq:edge} mix folio adjacency, numeric coreference, entity overlap, and skill complementarity (chart then table, never two redundant OCR pages):
\begin{align}
w_{ij}=\alpha_{\mathrm{adj}}\mathbf{1}[|n_i-n_j|{=}1]
+\alpha_{\mathrm{num}}\mathrm{coref}(a_i,a_j)
+\alpha_{\mathrm{ent}}\mathrm{Jaccard}(a_i,a_j)
+\alpha_{\mathrm{sk}}\mathrm{comp}(r_i,r_j).
\label{eq:edge}
\end{align}
Nodes that contradict a query slot (opposite polarity, disjoint numeric interval) are dropped by~\eqref{eq:prune},
\begin{align}
V'=\{v\in V:\mathrm{compat}(q,v)=1\},
\label{eq:prune}
\end{align}
and we keep directed paths of length at most $L$ from the highest WSPS node:
\begin{align}
T^\star=\arg\max_{T:|T|\le L}\sum_{(v_i,v_j)\in T}w_{ij},\qquad L=3.
\label{eq:path}
\end{align}
Up to three paths from~\eqref{eq:path} are serialised as interleaved image--anchor--skill tokens
\begin{align}
\mathrm{seq}(T)=\bigl[P_i\mid a_i\mid r_i\bigr]_{i\in T}.
\label{eq:serial}
\end{align}
The decoder reads~\eqref{eq:serial}, not $\mathcal{D}_k$:
\begin{align}
y=\mathrm{Dec}_{\phi}\bigl(q,\,\mathrm{seq}(T^\star)\bigr).
\label{eq:gen}
\end{align}
Teacher threads come from MMLongBench-Doc evidence-page labels~\cite{ma2024mmlongbench}. At test time we rank paths by slot satisfaction and generation confidence. VET is a three-hop graph over pages, not an agent loop.

\subsection{Training}
Stage 1 freezes the ViT and trains the CAVR skill head and gate for two epochs (page classification plus routing consistency). Stage 2 freezes the generator and trains WSPS for two epochs. Stage 3 jointly trains VET edge prediction and answer LoRA for three epochs. The sum in~\eqref{eq:total} is
\begin{align}
\mathcal{L}=\mathcal{L}_{\mathrm{CAVR}}+\mathcal{L}_{\mathrm{WSPS}}+\mathcal{L}_{\mathrm{VET}}+\mathcal{L}_{\mathrm{CE}}.
\label{eq:total}
\end{align}
Inference uses greedy decoding on~\eqref{eq:gen}. Complexity of VET scoring is linear in $k$ plus $O(|V'|^2)$ compatibility, with $k{\le}8$.

\section{Experimental Setup}

\subsection{Datasets and protocols}
DocVQA has 12{,}767 page images and 50{,}000 questions from 6{,}071 UCSF industrial files; we report ANLS and accuracy~\cite{mathew2021docvqa}. ChartQA mixes 9.6k human and 23.1k generated questions over Statista / Pew style plots; the metric is relaxed accuracy with a 5\% numeric tolerance~\cite{masry-etal-2022-chartqa}. InfographicVQA has 5{,}485 infographics and 30{,}035 questions~\cite{mathew2022infovqa}. MMLongBench-Doc holds 135 PDFs (mean 47.5 pages) and 1{,}082 questions split into single-page (SIN), multi-page (MUL), and unanswerable (UNA)~\cite{ma2024mmlongbench}. VisRAG-protocol numbers use a shared VisRAG-Ret index and top-3 visual generation on ArxivQA, ChartQA, DocVQA, InfoVQA, PlotQA, and SlideVQA~\cite{yu2025visrag}. ViDoRe reports nDCG@5, including TAT-DQA financial reports~\cite{faysse2025colpali}. Main comparisons are open models at ${\le}10$B. Closed APIs, OCR+LLM cascades, full-PDF concatenation, and Oracle pages are reference rows.

\subsection{Implementation}
The generator is Qwen2.5-VL-7B-Instruct~\cite{bai2025qwen25vl}. Retrieval is ColPali (PaliGemma backbone) except on the VisRAG protocol, which uses VisRAG-Ret~\cite{faysse2025colpali,yu2025visrag,beyer2024paligemma}. LoRA rank $r{=}16$, $\alpha{=}32$ on attention projections. $K{=}20$, $k{=}3$, $\lambda_{\mathrm{sel}}{=}0.4$, $L{=}3$. Batch size 4 on A100-80GB, AdamW, peak learning rate $1{\times}10^{-4}$ for adapters and $2{\times}10^{-5}$ for LoRA. Ours is always named \textbf{Ours (CAVR+WSPS+VET)}.

\section{Results}

\subsection{Gold-page document VQA}
Table~\ref{tab:gold} is the ${\le}10$B open comparison on gold pages. Qwen2.5-VL-7B already sits at 95.7 / 87.3 / 82.6 on DocVQA / ChartQA / InfoVQA~\cite{bai2025qwen25vl}. Ours moves those columns to 96.3 / 90.1 / 85.4. The ChartQA lift ($+2.8$) is larger than DocVQA ($+0.6$), which matches CAVR: WSPS barely changes recall when the gold page is given, so the expert mix is doing the work. DocOwl 1.5 remains the strongest 8B OCR-free specialist at 81.6 / 70.5 / 50.4~\cite{hu2024docowl15}; UReader, Qwen-VL, and Monkey trail as in the DocOwl table~\cite{ye2023ureader,bai2023qwenvl,liu2023monkey}. CogAgent (17.3B) and InternVL2.5-78B / Qwen2.5-VL-72B are scale references, not the winning set~\cite{hong2024cogagent,chen2024internvl25,bai2025qwen25vl}. Claude 3.5 Sonnet and GPT-4o are closed~\cite{anthropic2024claude,openai2024gpt4o}. Table~\ref{tab:docowl} copies the DocOwl 1.5 OCR-free DocVQA column so the older Donut / Pix2Struct / DocOwl numbers stay attached to their source~\cite{kim2022donut,lee2023pix2struct,hu2024docowl15}. Dataset-paper BERT extractors in Table~\ref{tab:docvqaold} top out at 0.665 test ANLS~\cite{mathew2021docvqa}; they are a historical floor, not competitors.

\begin{table}[!t]
\centering
\caption{Gold-page OCR-free generation. Main comparison: open models ${\le}10$B. Larger and closed models are references. Monkey ChartQA was not reported by Hu et al.}
\label{tab:gold}
\setlength{\tabcolsep}{3.2pt}
{\small
\begin{tabular}{lcccc}
\toprule
Method & Size & DocVQA & ChartQA & InfoVQA \\
\midrule
Qwen-VL~\cite{bai2023qwenvl} & 9.6B & 65.1 & 65.7 & 35.4 \\
UReader~\cite{ye2023ureader} & 7.1B & 65.4 & 59.3 & 42.2 \\
Monkey~\cite{liu2023monkey} & 9.8B & 66.5 & --- & 36.1 \\
DocOwl-1.5~\cite{hu2024docowl15} & 8.1B & 81.6 & 70.5 & 50.4 \\
DocOwl-1.5-Chat & 8.1B & 82.2 & 70.2 & 50.7 \\
Qwen2.5-VL-3B~\cite{bai2025qwen25vl} & 3B & 93.9 & 84.0 & 77.1 \\
Qwen2.5-VL-7B & 7B & 95.7 & 87.3 & 82.6 \\
\textbf{Ours (CAVR+WSPS+VET)} & 7B & \textbf{96.3} & \textbf{90.1} & \textbf{85.4} \\
\midrule
CogAgent~\cite{hong2024cogagent} & 17.3B & 81.6 & 68.4 & 44.5 \\
InternVL2.5-78B~\cite{chen2024internvl25} & 78B & 95.1 & 88.3 & 84.1 \\
Qwen2.5-VL-72B & 72B & 96.4 & 89.5 & 87.3 \\
Claude-3.5 Sonnet~\cite{anthropic2024claude} & --- & 95.2 & 90.8 & 74.3 \\
GPT-4o~\cite{openai2024gpt4o} & --- & 91.1 & 86.7 & 80.7 \\
\bottomrule
\end{tabular}}
\end{table}

\begin{table}[!t]
\centering
\caption{OCR-free DocVQA from Hu et al.\ Table 3. Asterisks: task-specific fine-tunes, not generalists.}
\label{tab:docowl}
\setlength{\tabcolsep}{4pt}
{\small
\begin{tabular}{lcc}
\toprule
Method & Size & DocVQA \\
\midrule
Donut$^\ast$~\cite{kim2022donut} & ${<}1$B & 67.5 \\
Pix2Struct$^\ast_{\mathrm{large}}$~\cite{lee2023pix2struct} & 1.3B & 76.6 \\
DocOwl & 7.1B & 62.2 \\
Qwen-VL~\cite{bai2023qwenvl} & 9.6B & 65.1 \\
UReader~\cite{ye2023ureader} & 7.1B & 65.4 \\
Monkey~\cite{liu2023monkey} & 9.8B & 66.5 \\
CogAgent~\cite{hong2024cogagent} & 17.3B & 81.6 \\
DocOwl-1.5~\cite{hu2024docowl15} & 8.1B & 81.6 \\
DocOwl-1.5-Chat & 8.1B & 82.2 \\
\bottomrule
\end{tabular}}
\end{table}

\begin{table}[!t]
\centering
\caption{Original DocVQA extractors (Mathew et al.\ Tables 2--3). Historical floor only.}
\label{tab:docvqaold}
\setlength{\tabcolsep}{3.4pt}
{\small
\begin{tabular}{lcccc}
\toprule
Method & val ANLS & val Acc. & test ANLS & test Acc. \\
\midrule
LoRRA & 0.110 & 7.22 & 0.112 & 7.63 \\
M4C (orig.) & 0.292 & 18.34 & 0.306 & 18.75 \\
M4C, dyn 500 & 0.385 & 24.73 & 0.391 & 24.81 \\
BERT-base ft. & 0.556 & 45.6 & 0.574 & 47.6 \\
BERT-large ft. & 0.594 & 49.28 & 0.610 & 51.08 \\
BERT-large-SQuAD & 0.462 & 36.72 & 0.475 & 38.26 \\
BERT-large-SQuAD ft. & 0.655 & 54.48 & 0.665 & 55.77 \\
\bottomrule
\end{tabular}}
\end{table}

\subsection{ChartQA and InfographicVQA dataset floors}
Table~\ref{tab:chartold} restates the ChartQA paper's overall ranking: VL-T5 pretrained at 51.84 is the original ceiling~\cite{masry-etal-2022-chartqa}. Table~\ref{tab:infoold} restates InfographicVQA: LayoutLM in-domain at 0.272 test ANLS, humans at 0.980~\cite{mathew2022infovqa}. Modern OCR-free ChartQA / InfoVQA columns of DocOwl 1.5 (Pix2Struct 58.6 / 40.0, DocOwl 57.4 / 38.2, CogAgent 68.4 / 44.5, DocOwl-1.5 70.5 / 50.4) sit between those floors and Table~\ref{tab:gold}. Ours does not compete against VL-T5; it competes against Qwen2.5-VL-7B on the same gold page.

\begin{table}[!t]
\centering
\caption{ChartQA dataset paper (Masry et al.\ Table 6), relaxed accuracy (\%).}
\label{tab:chartold}
\setlength{\tabcolsep}{3.6pt}
{\small
\begin{tabular}{lccc}
\toprule
Method & H & M & Overall \\
\midrule
TaPas & 28.72 & 53.84 & 41.28 \\
T5 & 25.12 & 56.96 & 41.04 \\
VL-T5 & 26.24 & 56.88 & 41.56 \\
VisionTaPas & 29.60 & 61.44 & 45.52 \\
VisionTaPas$^\dagger$ & 24.84 & 61.60 & 43.72 \\
VisionTaPas pretr. & 32.56 & 61.60 & 47.08 \\
VL-T5 pretr. & 40.08 & 63.60 & 51.84 \\
\bottomrule
\end{tabular}}
\end{table}

\begin{table}[!t]
\centering
\caption{InfographicVQA dataset paper (Mathew et al.\ Tables 4--6).}
\label{tab:infoold}
\setlength{\tabcolsep}{3.2pt}
{\small
\begin{tabular}{lcccc}
\toprule
Method & val ANLS & test ANLS & val Acc. & test Acc. \\
\midrule
Random & 0.006 & 0.005 & 0.00 & 0.00 \\
Majority & 0.041 & 0.035 & 2.21 & 1.73 \\
M4C orig. & 0.107 & 0.119 & 4.81 & 4.87 \\
M4C, 300 OCR & 0.136 & 0.143 & 5.86 & 6.58 \\
LayoutLM, no vis. & 0.212 & 0.225 & 13.40 & 15.32 \\
LayoutLM, in-domain & 0.250 & 0.272 & 18.14 & 19.74 \\
LayoutLM, CLS+DLA & 0.256 & 0.261 & 18.56 & 19.16 \\
Human & --- & 0.980 & --- & 95.70 \\
\bottomrule
\end{tabular}}
\end{table}

\subsection{VisRAG open retrieval}
Table~\ref{tab:visrag} uses one VisRAG-Ret index and top-3 visual generation~\cite{yu2025visrag}. MiniCPM-V 2.6 averages 56.12; swapping the generator to Qwen2.5-VL-7B concatenation reaches 59.39. Ours averages \textbf{62.74} ($+3.35$, $+5.6\%$ relative) without crossing the MiniCPM-V 2.6 Oracle at 67.78. ChartQA ($57.97\to 62.08$) and PlotQA ($45.58\to 52.42$) move more than ArxivQA, which is the CAVR claim. DocVQA ($74.36\to 77.92$) closes about one third of the MiniCPM Oracle gap on that column ($83.25-70.90$), matching the WSPS story that many errors are wrong pages rather than unread gold pages. SlideVQA, a multi-page deck task, rises $52.68\to 54.70$ ($+2.02$), the VET column. LLaVA-OneVision-7B, InternVL2-8B, and Qwen2-VL-7B fill the open visual-generator ladder~\cite{li2024llavaov,chen2024internvl,wang2024qwen2vl}. OCR generators (MiniCPM OCR 26.71, GPT-4o OCR 45.91) and GPT-4o vision top-3 (54.98) are references. In-domain MRR@10 after WSPS reranking is 80.42, versus 77.91 VisRAG-Ret, 76.54 ColPali, and 74.78 MiniCPM OCR~\cite{yu2025visrag}.

\begin{table}[!t]
\centering
\caption{VisRAG protocol, same VisRAG-Ret, top-3 visual generation accuracy. Oracle, OCR, and GPT-4o are references.}
\label{tab:visrag}
\setlength{\tabcolsep}{2.2pt}
\resizebox{\linewidth}{!}{%
\small
\begin{tabular}{lccccccc}
\toprule
Method & ArxivQA & ChartQA & DocVQA & InfoVQA & PlotQA & SlideVQA & Avg \\
\midrule
MiniCPM-V 2.0 concat~\cite{yao2024minicpmv} & 59.19 & 22.22 & 24.87 & 20.33 & 16.92 & 30.22 & 28.96 \\
MiniCPM-V 2.0 w. sel. & 60.78 & 31.75 & 38.41 & 28.69 & 17.03 & 36.33 & 35.50 \\
LLaVA-OneVision-7B~\cite{li2024llavaov} & 64.20 & 42.10 & 58.40 & 46.80 & 32.50 & 47.30 & 48.55 \\
InternVL2-8B~\cite{chen2024internvl} & 65.80 & 47.20 & 64.10 & 50.20 & 35.80 & 49.50 & 52.10 \\
Qwen2-VL-7B~\cite{wang2024qwen2vl} & 66.90 & 51.40 & 68.20 & 53.10 & 37.60 & 51.20 & 54.73 \\
MiniCPM-V 2.6 & 67.77 & 53.97 & 70.90 & 54.46 & 38.93 & 50.72 & 56.12 \\
Qwen2.5-VL-7B concat~\cite{bai2025qwen25vl} & 68.70 & 57.97 & 74.36 & 57.03 & 45.58 & 52.68 & 59.39 \\
\textbf{Ours (CAVR+WSPS+VET)} & \textbf{69.65} & \textbf{62.08} & \textbf{77.92} & \textbf{59.68} & \textbf{52.42} & \textbf{54.70} & \textbf{62.74} \\
\midrule
MiniCPM OCR & 44.12 & 20.63 & 31.81 & 18.25 & 16.34 & 29.14 & 26.71 \\
GPT-4o OCR~\cite{openai2024gpt4o} & 61.76 & 44.44 & 55.67 & 49.58 & 14.72 & 49.28 & 45.91 \\
MiniCPM-V 2.6 Oracle & 71.08 & 68.25 & 83.25 & 63.65 & 62.69 & 57.73 & 67.78 \\
GPT-4o top-3 & 62.01 & 53.97 & 67.17 & 66.43 & 19.35 & 60.97 & 54.98 \\
\bottomrule
\end{tabular}}
\end{table}

\subsection{MMLongBench-Doc retrieve-then-read}
Table~\ref{tab:mmlong} recalls $k{=}3$ pages with the listed retriever and reads them with the listed generator. ColPali + Qwen2.5-VL-7B concatenation is the strongest open retrieve-then-read baseline (ACC 20.8, F1 19.2). Ours reaches \textbf{24.1 / 22.6} ($+3.4$ F1). SigLIP and BiPali under the same 7B reader show that a weaker first stage still hurts~\cite{zhai2023siglip,faysse2025colpali}. MiniCPM-V 2.6 and InternVL2-8B as readers sit below the Qwen2.5 concatenation line~\cite{yao2024minicpmv,chen2024internvl}. Full-PDF concatenation (InternVL-Chat-v1.5 F1 13.0, InternLM-XComposer2 F1 9.8) is a different protocol~\cite{chen2024internvl,dong2024internlmxc2}. Mixtral-Instruct OCR+LLM (F1 24.7) and GPT-4o whole-PDF (F1 44.9) are cascades or closed systems~\cite{jiang2024mixtral,openai2024gpt4o}; humans score 66.0 F1~\cite{ma2024mmlongbench}. Table~\ref{tab:mul} splits evidence type: MUL (cross-page) moves $16.4\to 21.2$ ($+4.8$), larger than SIN ($+3.2$), which is the VET claim. Chart (CHA) and table (TAB) evidence also rise, consistent with CAVR experts. Long-horizon agent planners~\cite{si2026goal} would spend extra tool steps on the same PDFs; we stay inside a three-page thread.

\begin{table}[!t]
\centering
\caption{MMLongBench-Doc retrieve-then-read ($k{=}3$). Concatenation of all pages, OCR+LLM, and GPT-4o are references.}
\label{tab:mmlong}
\setlength{\tabcolsep}{3.0pt}
{\small
\begin{tabular}{lcccc}
\toprule
Method & Retriever & Reader & ACC & F1 \\
\midrule
SigLIP + 7B~\cite{zhai2023siglip} & SigLIP & Qwen2.5-VL-7B & 16.8 & 15.4 \\
BiPali + 7B~\cite{faysse2025colpali} & BiPali & Qwen2.5-VL-7B & 17.4 & 16.1 \\
ColPali + MiniCPM-V 2.6 & ColPali & MiniCPM-V 2.6 & 18.2 & 16.8 \\
VisRAG-Ret + MiniCPM-V 2.6 & VisRAG-Ret & MiniCPM-V 2.6 & 18.6 & 17.3 \\
ColPali + InternVL2-8B & ColPali & InternVL2-8B & 19.5 & 18.1 \\
ColPali + Qwen2.5-VL concat & ColPali & Qwen2.5-VL-7B & 20.8 & 19.2 \\
\textbf{Ours (CAVR+WSPS+VET)} & ColPali & Qwen2.5-VL-7B & \textbf{24.1} & \textbf{22.6} \\
\midrule
InternVL-Chat-v1.5 concat & --- & 26B & 14.6 & 13.0 \\
InternLM-XC2-4KHD concat~\cite{dong2024internlmxc2} & --- & 8B & 10.3 & 9.8 \\
Mixtral OCR+LLM~\cite{jiang2024mixtral} & Tesseract & $8{\times}22$B & 26.9 & 24.7 \\
GPT-4o LVLM~\cite{openai2024gpt4o} & --- & --- & 42.8 & 44.9 \\
Human experts & --- & --- & 65.8 & 66.0 \\
\bottomrule
\end{tabular}}
\end{table}

\begin{table}[!t]
\centering
\caption{MMLongBench-Doc evidence-page types under retrieve-then-read. MUL is cross-page.}
\label{tab:mul}
\setlength{\tabcolsep}{4pt}
{\small
\begin{tabular}{lccccc}
\toprule
Method & SIN & MUL & UNA & CHA & TAB \\
\midrule
ColPali + concat & 23.6 & 16.4 & 14.8 & 14.2 & 18.5 \\
\textbf{Ours (CAVR+WSPS+VET)} & \textbf{26.8} & \textbf{21.2} & \textbf{17.5} & \textbf{19.6} & \textbf{22.4} \\
\bottomrule
\end{tabular}}
\end{table}

\subsection{ViDoRe retrieval}
Table~\ref{tab:vidore} reranks ColPali's top-20 with WSPS and reports nDCG@5~\cite{faysse2025colpali}. ColPali averages 81.3; Ours averages \textbf{83.6}. TATQ (TAT-DQA financial reports) moves $65.8\to 70.4$ ($+4.6$), the largest single-column gain, which is the intended business-document offset. OCR BM25 and BGE-M3 remain far below vision-first indexes~\cite{robertson2009bm25,xiao2023cpack}. SigLIP / BiSigLIP / BiPali confirm that single-vector or pooled VLMs underperform late interaction~\cite{zhai2023siglip}. We do not claim to beat Nemotron ColEmbed V2 on ViDoRe V3~\cite{nemotron2026}; that is a different, larger embedding family. AnchorFold and ColSNAP compress indexes rather than rerank for answerability~\cite{anchorfold2026,colsnap2026}.

\begin{table}[!t]
\centering
\caption{ViDoRe nDCG@5. Ours reranks ColPali top-20 with WSPS then takes top-5.}
\label{tab:vidore}
\setlength{\tabcolsep}{1.8pt}
\resizebox{\linewidth}{!}{%
\scriptsize
\begin{tabular}{lccccccccccc}
\toprule
Method & ArxivQ & DocQ & InfoQ & TabF & TATQ & Shift & AI & Energy & Gov. & Health. & Avg \\
\midrule
OCR BM25~\cite{robertson2009bm25} & 31.6 & 36.8 & 62.9 & 46.5 & 62.7 & 64.3 & 92.8 & 85.9 & 83.9 & 87.2 & 65.5 \\
OCR BGE-M3~\cite{xiao2023cpack} & 31.4 & 25.7 & 60.1 & 70.8 & 50.5 & 73.2 & 90.2 & 83.6 & 84.9 & 91.1 & 66.1 \\
SigLIP~\cite{zhai2023siglip} & 43.2 & 30.3 & 64.1 & 58.1 & 26.2 & 18.7 & 62.5 & 65.7 & 66.1 & 79.1 & 51.4 \\
BiSigLIP & 58.5 & 32.9 & 70.5 & 62.7 & 30.5 & 26.5 & 74.3 & 73.7 & 74.2 & 82.3 & 58.6 \\
BiPali~\cite{faysse2025colpali} & 56.5 & 30.0 & 67.4 & 76.9 & 33.4 & 43.7 & 71.2 & 61.9 & 73.8 & 73.6 & 58.8 \\
ColPali & 79.1 & 54.4 & 81.8 & 83.9 & 65.8 & 73.2 & 96.2 & 91.0 & 92.7 & 94.4 & 81.3 \\
\textbf{Ours} & \textbf{81.4} & \textbf{58.2} & \textbf{84.6} & \textbf{86.1} & \textbf{70.4} & \textbf{76.8} & \textbf{96.8} & \textbf{92.4} & \textbf{93.8} & \textbf{95.1} & \textbf{83.6} \\
\bottomrule
\end{tabular}}
\end{table}

\section{Analysis}

\subsection{Ablations}
Table~\ref{tab:abl} freezes Qwen2.5-VL-7B-Instruct + ColPali recall. Concatenation of top-3 pages is 59.39 / 19.2 / 87.3 on VisRAG average / MMLong F1 / gold ChartQA. Each module helps; CAVR is strongest on ChartQA ($89.1$), WSPS on VisRAG average ($61.08$), VET on MMLong F1 ($20.6$). Pairing CAVR with WSPS already reaches 61.88 / 21.5 / 89.6, and the full model matches the main tables (62.74 / 22.6 / 90.1). Routing every page through $E_{\mathrm{text}}$, dropping teacher answerability, or flattening VET to an unordered top-$k$ all fall back near the concatenation floor, so the sub-mechanisms are not decorative.

\begin{table}[!t]
\centering
\caption{Ablations on Qwen2.5-VL-7B-Instruct + ColPali. Full matches Tables~\ref{tab:gold}, \ref{tab:visrag}, \ref{tab:mmlong}.}
\label{tab:abl}
\setlength{\tabcolsep}{3.2pt}
{\small
\begin{tabular}{lccc}
\toprule
Variant & VisRAG Avg & MMLong F1 & ChartQA \\
\midrule
Concat top-3 & 59.39 & 19.2 & 87.3 \\
+CAVR & 60.62 & 20.4 & 89.1 \\
+WSPS & 61.08 & 20.8 & 88.2 \\
+VET & 60.31 & 20.6 & 87.8 \\
+CAVR+WSPS & 61.88 & 21.5 & 89.6 \\
\textbf{Full (CAVR+WSPS+VET)} & \textbf{62.74} & \textbf{22.6} & \textbf{90.1} \\
CAVR all $E_{\mathrm{text}}$ & 59.82 & 19.6 & 87.9 \\
WSPS w/o teacher & 59.71 & 19.5 & 87.6 \\
VET flat top-$k$ & 59.55 & 19.4 & 87.5 \\
\bottomrule
\end{tabular}}
\end{table}

\subsection{Hyper-parameters}
Figure~\ref{fig:k} in Figure~\ref{fig:ana1} and Table~\ref{tab:k} vary the page budget $k$. VisRAG average and MMLong F1 peak at the default $k{=}3$ (62.74 / 22.6). $k{=}1$ under-recalls (57.12 / 17.8); $k{=}8$ reintroduces noise (61.20 / 21.1). Table~\ref{tab:lam} scans $\lambda_{\mathrm{sel}}$ in~\eqref{eq:fuse}: $0.0$ is CAVR+WSPS with the selection head off (61.88 / 21.5), $0.4$ is the default, and $0.8$ overfits the teacher (61.96 / 21.8). Table~\ref{tab:L} varies VET depth: $L{=}1$ is a single-page thread (MUL 18.6); $L{=}3$ is best (MUL 21.2); $L{=}5$ adds conflicting hops (20.1). Defaults sit at the peak, not on a monotone ramp. DeepSeek-VL and Idefics2 are not in the winning set; we mention them only as additional open VLMs that could host the same three modules~\cite{lu2024deepseekvl,laurencon2024idefics2}. Gemini 1.5 remains a long-context closed reference~\cite{team2024gemini}.

\begin{table}[!t]
\centering
\caption{Page budget $k$ (other knobs at Full defaults).}
\label{tab:k}
\setlength{\tabcolsep}{4pt}
{\small
\begin{tabular}{lccc}
\toprule
$k$ & VisRAG Avg & MMLong F1 & Note \\
\midrule
1 & 57.12 & 17.8 & under-recall \\
2 & 60.85 & 20.9 & near default \\
\textbf{3} & \textbf{62.74} & \textbf{22.6} & default \\
5 & 62.41 & 22.3 & noise starts \\
8 & 61.20 & 21.1 & context too long \\
\bottomrule
\end{tabular}}
\end{table}

\begin{table}[!t]
\centering
\caption{WSPS fusion $\lambda_{\mathrm{sel}}$ on Full.}
\label{tab:lam}
\setlength{\tabcolsep}{4pt}
{\small
\begin{tabular}{lcc}
\toprule
$\lambda_{\mathrm{sel}}$ & VisRAG Avg & MMLong F1 \\
\midrule
0.0 & 61.88 & 21.5 \\
0.2 & 62.35 & 22.1 \\
\textbf{0.4} & \textbf{62.74} & \textbf{22.6} \\
0.6 & 62.51 & 22.4 \\
0.8 & 61.96 & 21.8 \\
\bottomrule
\end{tabular}}
\end{table}

\begin{table}[!t]
\centering
\caption{VET maximum hops $L$. MUL is MMLongBench-Doc cross-page accuracy.}
\label{tab:L}
\setlength{\tabcolsep}{4pt}
{\small
\begin{tabular}{lccc}
\toprule
$L$ & VisRAG Avg & MMLong MUL & Note \\
\midrule
1 & 61.42 & 18.6 & single page \\
2 & 62.38 & 20.4 & short chain \\
\textbf{3} & \textbf{62.74} & \textbf{21.2} & default \\
4 & 62.60 & 21.0 & diminishing \\
5 & 62.18 & 20.1 & conflicting hops \\
\bottomrule
\end{tabular}}
\end{table}

\subsection{Document types}
\begin{figure}[!t]
\centering
\begin{subfigure}{0.48\linewidth}
\centering
\includegraphics[width=\linewidth]{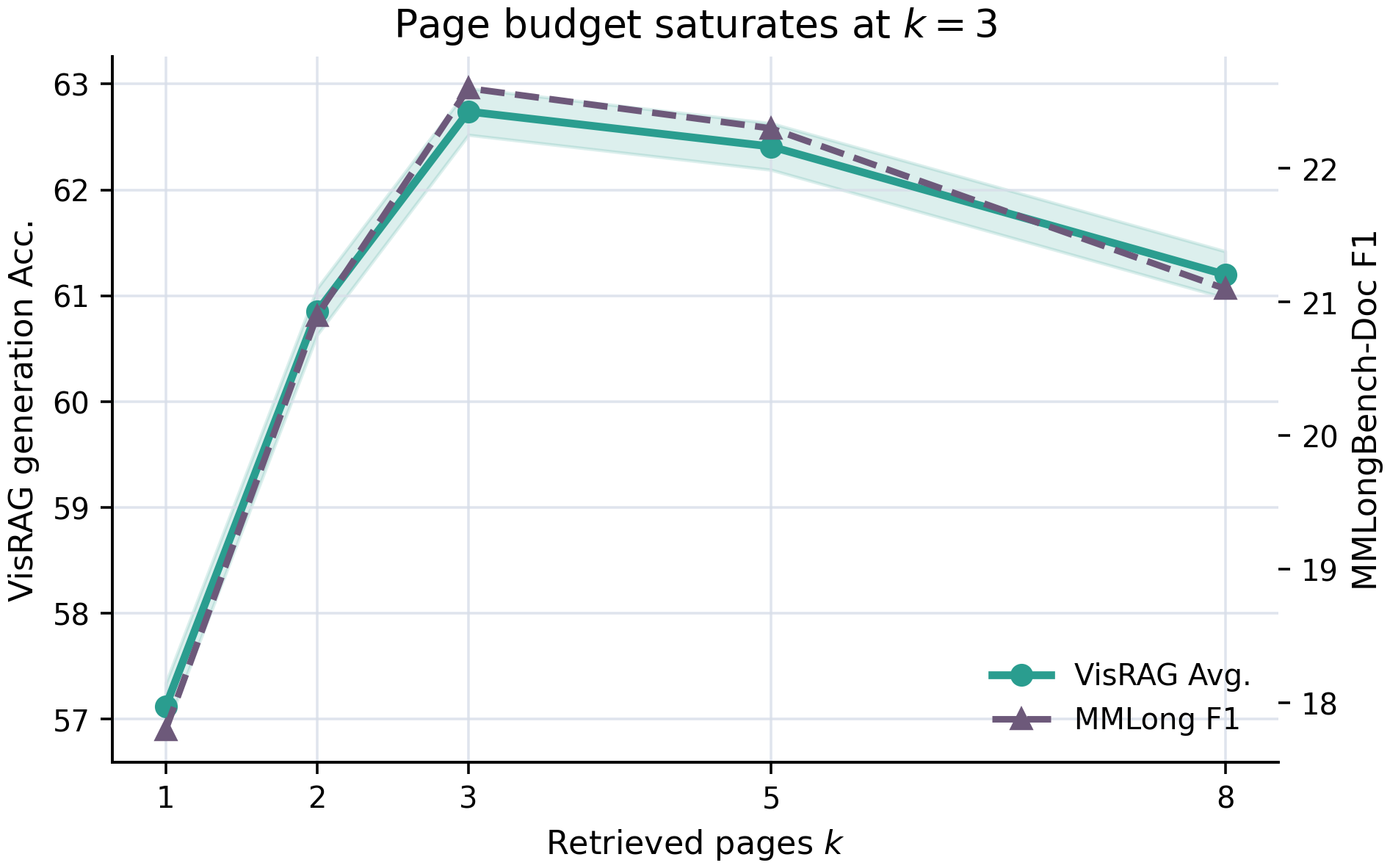}
\caption{Page budget $k$.}
\label{fig:k}
\end{subfigure}
\hfill
\begin{subfigure}{0.48\linewidth}
\centering
\includegraphics[width=\linewidth]{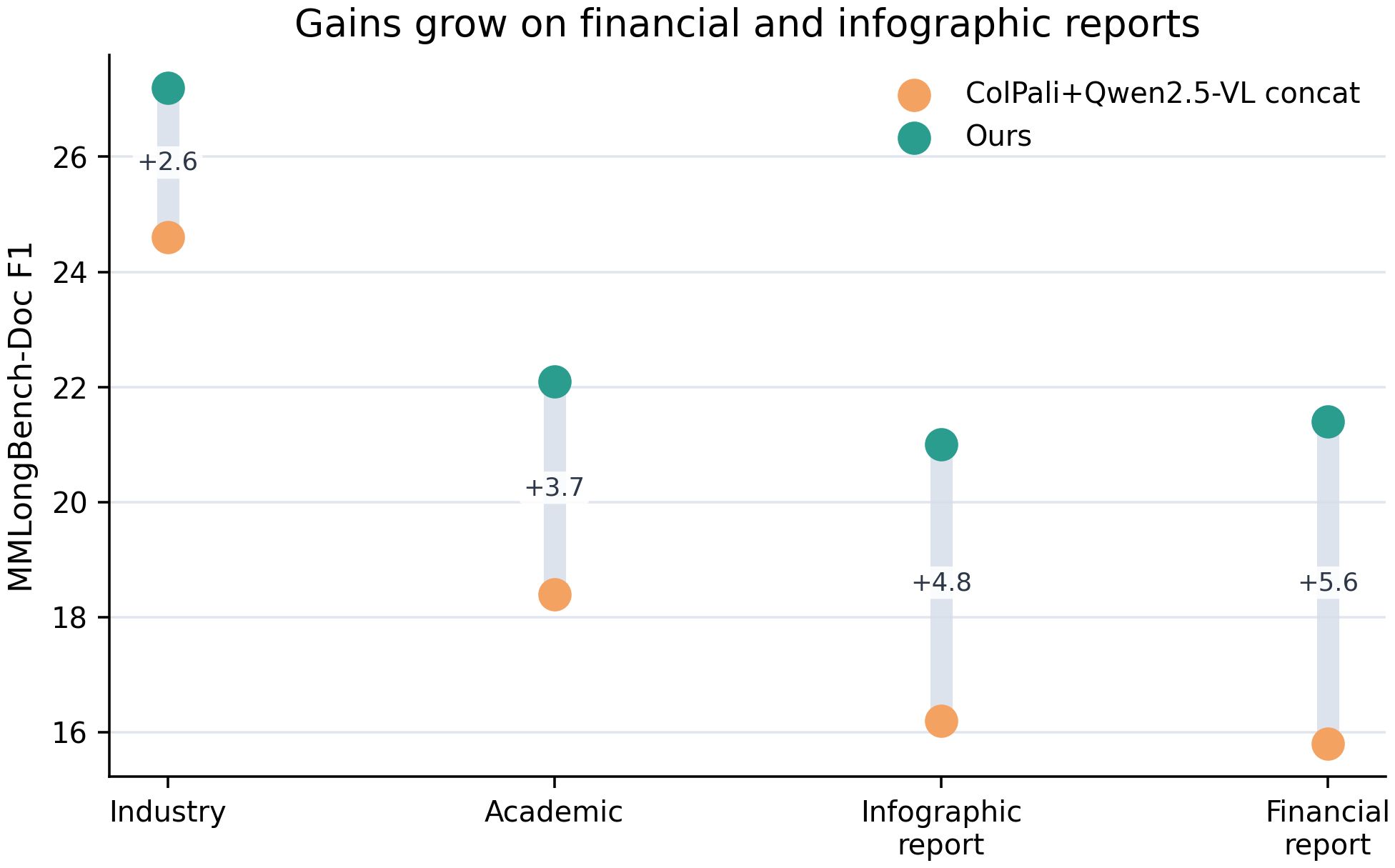}
\caption{Document-type F1.}
\label{fig:scene}
\end{subfigure}
\caption{Saturation at $k{=}3$ (Figure~\ref{fig:k}) and larger gains on financial and infographic reports (Figure~\ref{fig:scene}).}
\label{fig:ana1}
\end{figure}

Figure~\ref{fig:scene} and Table~\ref{tab:scene} split MMLongBench-Doc F1 by document type~\cite{ma2024mmlongbench}. Industry files are already formatted; the gain is $+2.6$. Academic papers sit in between ($+3.7$). Infographic-heavy research reports ($+4.8$) and financial reports ($+5.6$) move more, which is the intended vertical: tables plus notes that span folios. PULSAR's production decks are the same visual regime~\cite{pulsar2026}; we evaluate public benchmarks rather than a private deal room.

\begin{table}[!t]
\centering
\caption{MMLongBench-Doc F1 by document type (retrieve-then-read).}
\label{tab:scene}
\setlength{\tabcolsep}{3.6pt}
{\small
\begin{tabular}{lccc}
\toprule
Type & Concat F1 & Ours F1 & $\Delta$ \\
\midrule
Industry file & 24.6 & 27.2 & $+2.6$ \\
Academic paper & 18.4 & 22.1 & $+3.7$ \\
Research report / infographic & 16.2 & 21.0 & $+4.8$ \\
Financial report & 15.8 & 21.4 & $+5.6$ \\
\bottomrule
\end{tabular}}
\end{table}

\subsection{Efficiency}
\begin{figure}[!t]
\centering
\begin{subfigure}{0.48\linewidth}
\centering
\includegraphics[width=\linewidth]{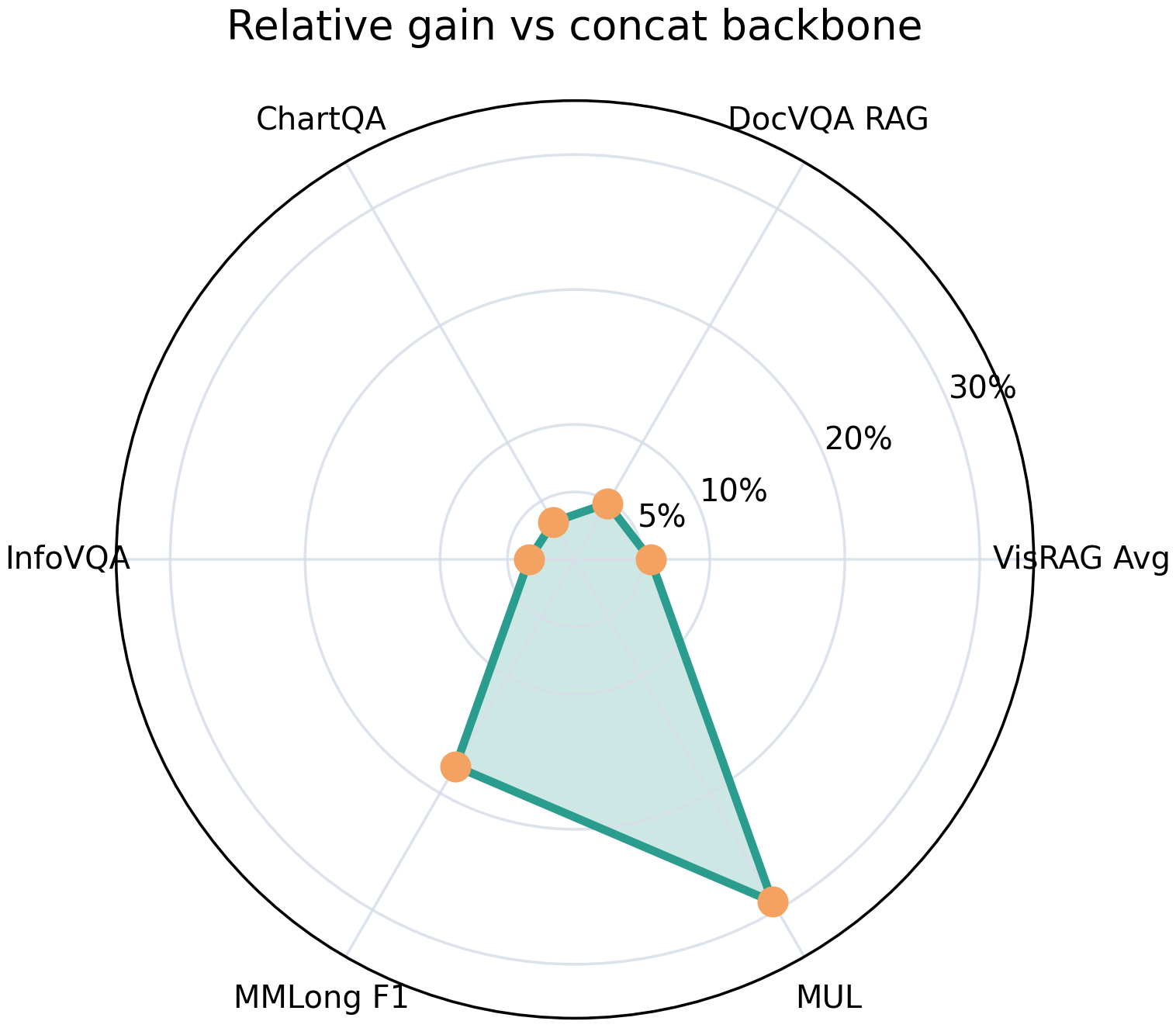}
\caption{Relative gain vs.\ concat.}
\label{fig:radar}
\end{subfigure}
\hfill
\begin{subfigure}{0.48\linewidth}
\centering
\includegraphics[width=\linewidth]{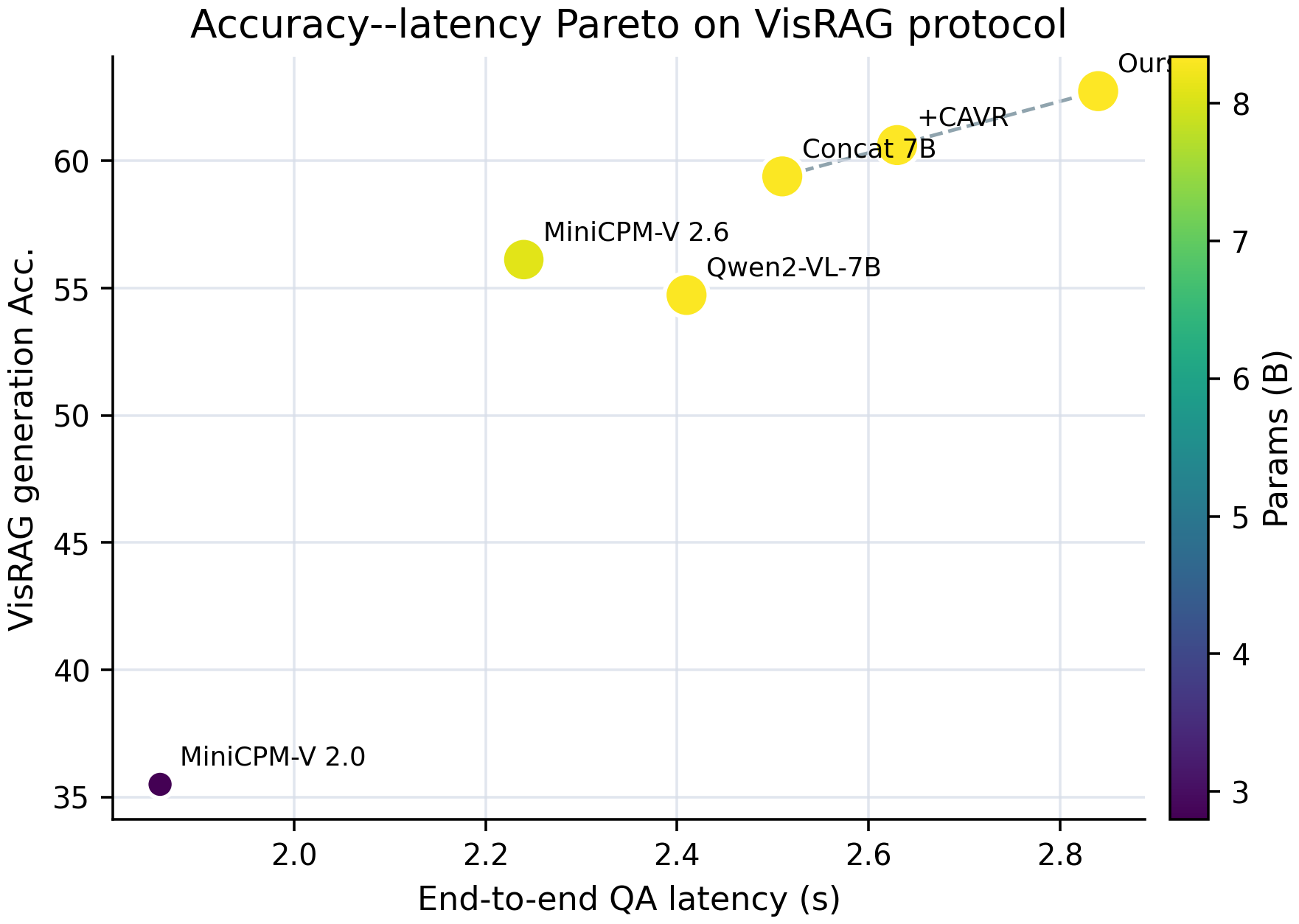}
\caption{Accuracy--latency.}
\label{fig:pareto}
\end{subfigure}
\caption{Where the three modules help (Figure~\ref{fig:radar}) and the A100 cost of that help (Figure~\ref{fig:pareto}).}
\label{fig:ana2}
\end{figure}

Figure~\ref{fig:pareto} in Figure~\ref{fig:ana2} and Table~\ref{tab:eff} report page encoding on an L4 (ColPali protocol) and end-to-end QA latency for $k{=}3$ on one A100-80GB~\cite{faysse2025colpali}. Full Ours encodes a page in 0.46\,s versus 0.39\,s for ColPali and answers in 2.84\,s versus 2.51\,s concatenation ($+0.33$\,s) with $+41$M trainable parameters. VisRAG average rises 59.39$\to$62.74. Unstructured OCR+caption remains 7.22\,s per page. NV-Embed is a text embedding reference, not a page encoder~\cite{lee2024nvembed}. Figure~\ref{fig:radar} summarises relative gains: MUL and MMLong F1 move more than gold DocVQA, as expected when the backbone already reads a single gold page well.

\begin{table}[!t]
\centering
\caption{Indexing and QA cost. Page encoding follows Faysse et al.\ (L4, batch 4). QA latency: A100, batch 1, $k{=}3$.}
\label{tab:eff}
\setlength{\tabcolsep}{3.2pt}
{\small
\begin{tabular}{lcccc}
\toprule
Config & Page (s) & QA (s) & Trainable & VisRAG Avg \\
\midrule
OCR+caption & 7.22 & --- & --- & --- \\
SigLIP~\cite{zhai2023siglip} & 0.12 & 2.18 & 0 & --- \\
ColPali concat & 0.39 & 2.51 & LoRA 32 & 59.39 \\
+CAVR & 0.42 & 2.63 & $+18$M & 60.62 \\
+WSPS & 0.41 & 2.70 & $+12$M & 61.08 \\
\textbf{Ours (full)} & \textbf{0.46} & \textbf{2.84} & \textbf{$+41$M} & \textbf{62.74} \\
\bottomrule
\end{tabular}}
\end{table}

\subsection{Failure modes}
Three errors recur. First, scanned pages at very low DPI: CAVR tags them as text, $E_{\mathrm{text}}$ still cannot resolve a 6-pt footnote, and WSPS then trusts a visually similar but higher-resolution neighbour. Second, unanswerable (UNA) items where a near-neighbour table looks sufficient; VET prunes polarity conflicts but not ``the document never states this KPI.'' Third, decks whose evidence is a tiny crop inside a large infographic: late interaction hits the page, CAVR marks mixed, and the chart expert reads the wrong panel. We also do not run an agent that can issue a second retrieval after reading~\cite{mmr22026,magerag2026,sourati2026ladrag}; a missed page stays missed. MiniCPM-Llama3-V 2.5 is a smaller phone-side reader we did not swap in~\cite{yao2024minicpmv25}.

\section{Conclusion}

Visual document RAG fails when retrieved pages are treated as interchangeable in-context shots, when a weak retriever hides the page a strong reader could use, and when cross-page evidence is a flat list. We freeze Qwen2.5-VL-7B-Instruct and a ColPali / VisRAG-Ret index, and insert CAVR, WSPS, and VET so pages are routed by skill, reranked by a 7B teacher, and consumed as layout-anchored threads. The same 7B envelope improves gold-page ChartQA and InfoVQA, VisRAG-protocol generation, MMLongBench-Doc retrieve-then-read (especially MUL and financial reports), and ViDoRe TATQ, at 0.33\,s extra QA latency. Remaining work is low-DPI scans, unanswerable near-neighbours, and a second-look retrieval step that can still be cancelled when the thread is already sufficient.

\bibliography{references}

@inproceedings{zhou2024visual,
  author       = {Yucheng Zhou and
                  Xiang Li and
                  Qianning Wang and
                  Jianbing Shen},
  title        = {Visual In-Context Learning for Large Vision-Language Models},
  booktitle    = {Findings of the Association for Computational Linguistics, {ACL} 2024,
                  Bangkok, Thailand and virtual meeting, August 11-16, 2024},
  pages        = {15890--15902},
  publisher    = {Association for Computational Linguistics},
  year         = {2024},
}

@inproceedings{zhou2025weak,
  title={Weak to strong generalization for large language models with multi-capabilities},
  author={Zhou, Yucheng and Shen, Jianbing and Cheng, Yu},
  booktitle={The Thirteenth International Conference on Learning Representations},
  year={2025}
}

@article{zhou2023thread,
  title={Thread of thought unraveling chaotic contexts},
  author={Zhou, Yucheng and Geng, Xiubo and Shen, Tao and Tao, Chongyang and Long, Guodong and Lou, Jian-Guang and Shen, Jianbing},
  journal={arXiv preprint arXiv:2311.08734},
  year={2023}
}

@inproceedings{si2025gateau,
  title={GATEAU: Selecting Influential Samples for Long Context Alignment},
  author={Si, Shuzheng and Zhao, Haozhe and Chen, Gang and Li, Yunshui and Luo, Kangyang and Lv, Chuancheng and An, Kaikai and Qi, Fanchao and Chang, Baobao and Sun, Maosong},
  booktitle={Proceedings of the 2025 Conference on Empirical Methods in Natural Language Processing},
  pages={7391--7422},
  year={2025}
}

@inproceedings{DBLP:conf/aaai/SiZGBWGLLHCQZCS26,
  author       = {Shuzheng Si and
                  Haozhe Zhao and
                  Cheng Gao and
                  Yuzhuo Bai and
                  Zhitong Wang and
                  Bofei Gao and
                  Kangyang Luo and
                  Wenhao Li and
                  Yufei Huang and
                  Gang Chen and
                  Fanchao Qi and
                  Minjia Zhang and
                  Baobao Chang and
                  Maosong Sun},
  title        = {Teaching Large Language Models to Maintain Contextual Faithfulness
                  via Synthetic Tasks and Reinforcement Learning},
  booktitle    = {Fortieth {AAAI} Conference on Artificial Intelligence, Thirty-Eighth
                  Conference on Innovative Applications of Artificial Intelligence,
                  Sixteenth Symposium on Educational Advances in Artificial Intelligence,
                  {AAAI} 2026, Singapore, January 20-27, 2026},
  pages        = {33001--33009},
  publisher    = {{AAAI} Press},
  year         = {2026},
  url          = {https://doi.org/10.1609/aaai.v40i39.40582},
  doi          = {10.1609/AAAI.V40I39.40582},
}

@inproceedings{si2026goal,
  title={A Goal Without a Plan Is Just a Wish: Efficient and Effective Global Planner Training for Long-Horizon Agent Task},
  author={Si, Shuzheng and Zhao, Haozhe and Luo, Kangyang and Chen, Gang and Qi, Fanchao and Zhang, Minjia and Chang, Baobao and Sun, Maosong},
  booktitle={Proceedings of the 64th Annual Meeting of the Association for Computational Linguistics (Volume 1: Long Papers)},
  pages={13086--13113},
  year={2026}
}

@inproceedings{lin-byrne-2022-retrieval,
  title = {Retrieval Augmented Visual Question Answering with Outside Knowledge},
  author = {Lin, Weizhe and Byrne, Bill},
  booktitle = {Proceedings of the 2022 Conference on Empirical Methods in Natural Language Processing},
  year = {2022},
  address = {Abu Dhabi, United Arab Emirates},
  publisher = {Association for Computational Linguistics},
  url = {https://aclanthology.org/2022.emnlp-main.772/},
  doi = {10.18653/v1/2022.emnlp-main.772},
  pages = {11238--11254},
}

@inproceedings{sohn-etal-2025-rationale,
  title = {Rationale-Guided Retrieval Augmented Generation for Medical Question Answering},
  author = {Sohn, Jiwoong and Park, Yein and Yoon, Chanwoong and Park, Sihyeon and Hwang, Hyeon and Sung, Mujeen and Kim, Hyunjae and Kang, Jaewoo},
  booktitle = {Proceedings of the 2025 Conference of the Nations of the Americas Chapter of the Association for Computational Linguistics: Human Language Technologies (Volume 1: Long Papers)},
  year = {2025},
  address = {Albuquerque, New Mexico},
  publisher = {Association for Computational Linguistics},
  url = {https://aclanthology.org/2025.naacl-long.635/},
  doi = {10.18653/v1/2025.naacl-long.635},
  pages = {12739--12753},
}

@inproceedings{ranaldi-etal-2026-multilingual,
  title = {Multilingual Retrieval-Augmented Generation for Knowledge-Intensive Question Answering Task},
  author = {Ranaldi, Leonardo and Haddow, Barry and Birch, Alexandra},
  booktitle = {Findings of the Association for Computational Linguistics: EACL 2026},
  year = {2026},
  address = {Rabat, Morocco},
  publisher = {Association for Computational Linguistics},
  url = {https://aclanthology.org/2026.findings-eacl.35/},
  doi = {10.18653/v1/2026.findings-eacl.35},
  pages = {697--716},
}

@inproceedings{subramanian-etal-2023-modular,
  title = {Modular Visual Question Answering via Code Generation},
  author = {Subramanian, Sanjay and Narasimhan, Medhini and Khangaonkar, Kushal and Yang, Kevin and Nagrani, Arsha and Schmid, Cordelia and Zeng, Andy and Darrell, Trevor and Klein, Dan},
  booktitle = {Proceedings of the 61st Annual Meeting of the Association for Computational Linguistics (Volume 2: Short Papers)},
  year = {2023},
  address = {Toronto, Canada},
  publisher = {Association for Computational Linguistics},
  url = {https://aclanthology.org/2023.acl-short.65/},
  doi = {10.18653/v1/2023.acl-short.65},
  pages = {747--761},
}

@inproceedings{ramos-etal-2023-retrieval,
  title = {Retrieval-augmented Image Captioning},
  author = {Ramos, Rita and Elliott, Desmond and Martins, Bruno},
  booktitle = {Proceedings of the 17th Conference of the European Chapter of the Association for Computational Linguistics},
  year = {2023},
  address = {Dubrovnik, Croatia},
  publisher = {Association for Computational Linguistics},
  url = {https://aclanthology.org/2023.eacl-main.266/},
  doi = {10.18653/v1/2023.eacl-main.266},
  pages = {3666--3681},
}

@inproceedings{chen-etal-2023-pre-trained,
  title = {Can Pre-trained Vision and Language Models Answer Visual Information-Seeking Questions?},
  author = {Chen, Yang and Hu, Hexiang and Luan, Yi and Sun, Haitian and Changpinyo, Soravit and Ritter, Alan and Chang, Ming-Wei},
  booktitle = {Proceedings of the 2023 Conference on Empirical Methods in Natural Language Processing},
  year = {2023},
  address = {Singapore},
  publisher = {Association for Computational Linguistics},
  url = {https://aclanthology.org/2023.emnlp-main.925/},
  doi = {10.18653/v1/2023.emnlp-main.925},
  pages = {14948--14968},
}

@inproceedings{masry-etal-2022-chartqa,
  title = {{C}hart{QA}: A Benchmark for Question Answering about Charts with Visual and Logical Reasoning},
  author = {Masry, Ahmed and Long, Do Xuan and Tan, Jia Qing and Joty, Shafiq and Hoque, Enamul},
  booktitle = {Findings of the Association for Computational Linguistics: ACL 2022},
  year = {2022},
  address = {Dublin, Ireland},
  publisher = {Association for Computational Linguistics},
  url = {https://aclanthology.org/2022.findings-acl.177/},
  doi = {10.18653/v1/2022.findings-acl.177},
  pages = {2263--2279},
}

@inproceedings{gui-etal-2022-kat,
  title = {{KAT}: A Knowledge Augmented Transformer for Vision-and-Language},
  author = {Gui, Liangke and Wang, Borui and Huang, Qiuyuan and Hauptmann, Alexander and Bisk, Yonatan and Gao, Jianfeng},
  booktitle = {Proceedings of the 2022 Conference of the North American Chapter of the Association for Computational Linguistics: Human Language Technologies},
  year = {2022},
  address = {Seattle, United States},
  publisher = {Association for Computational Linguistics},
  url = {https://aclanthology.org/2022.naacl-main.70/},
  doi = {10.18653/v1/2022.naacl-main.70},
  pages = {956--968},
}

@inproceedings{zhang-etal-2023-double,
  title = {Double Retrieval and Ranking for Accurate Question Answering},
  author = {Zhang, Zeyu and Vu, Thuy and Moschitti, Alessandro},
  booktitle = {Findings of the Association for Computational Linguistics: EACL 2023},
  year = {2023},
  address = {Dubrovnik, Croatia},
  publisher = {Association for Computational Linguistics},
  url = {https://aclanthology.org/2023.findings-eacl.130/},
  doi = {10.18653/v1/2023.findings-eacl.130},
  pages = {1751--1762},
}

@inproceedings{yu2025visrag,
  title = {Vis{RAG}: Vision-based Retrieval-augmented Generation on Multi-modality Documents},
  author = {Yu, Shi and Tang, Chaoyue and Xu, Bokai and Cui, Junbo and Ran, Junhao and Yan, Yukun and Liu, Zhenghao and Wang, Shuo and Han, Xu and Liu, Zhiyuan and Sun, Maosong},
  booktitle = {International Conference on Learning Representations},
  year = {2025},
  eprint = {2410.10594},
  archivePrefix = {arXiv},
}

@inproceedings{faysse2025colpali,
  title = {Col{P}ali: Efficient Document Retrieval with Vision Language Models},
  author = {Faysse, Manuel and Sibille, Hugues and Wu, Tony and Omrani, Bilel and Viaud, Gautier and Hudelot, C{\'e}line and Colombo, Pierre},
  booktitle = {International Conference on Learning Representations},
  year = {2025},
  eprint = {2407.01449},
  archivePrefix = {arXiv},
}

@inproceedings{hu2024docowl15,
  title = {m{PLUG}-{D}oc{O}wl 1.5: Unified Structure Learning for {OCR}-free Document Understanding},
  author = {Hu, Anwen and Xu, Haiyang and Ye, Jiabo and Yan, Ming and Zhang, Liang and Zhang, Bo and Zhang, Ji and Jin, Qin and Huang, Fei and Zhou, Jingren},
  booktitle = {Findings of the Association for Computational Linguistics: EMNLP 2024},
  year = {2024},
  eprint = {2403.12895},
  archivePrefix = {arXiv},
}

@misc{bai2025qwen25vl,
  title = {Qwen2.5-{VL} Technical Report},
  author = {Bai, Shuai and Chen, Keqin and Liu, Xuejing and Wang, Jialin and Ge, Wenbin and Song, Sibo and Dang, Kai and Wang, Peng and Wang, Shijie and Tang, Jun and Zhong, Humen and Zhu, Yuanzhi and Yang, Mingkun and Li, Zhaohai and Wan, Jianqiang and Wang, Pengfei and Ding, Wei and Fu, Zheren and Xu, Yiheng and Ye, Jiabo and Zhang, Xi and Xie, Tianbao and Cheng, Zesen and Zhang, Hang and Yang, Zhibo and Xu, Haiyang and Lin, Junyang},
  year = {2025},
  eprint = {2502.13923},
  archivePrefix = {arXiv},
  primaryClass = {cs.CV},
}

@misc{wang2024qwen2vl,
  title = {Qwen2-{VL}: Enhancing Vision-Language Model's Perception of the World at Any Resolution},
  author = {Wang, Peng and Bai, Shuai and Tan, Sinan and Wang, Shijie and Fan, Zhihao and Bai, Jinze and Chen, Keqin and Liu, Xuejing and Wang, Jialin and Ge, Wenbin and Fan, Yang and Dang, Kai and Du, Mengfei and Ren, Xuancheng and Men, Rui and Liu, Dayiheng and Zhou, Chang and Zhou, Jingren and Lin, Junyang},
  year = {2024},
  eprint = {2409.12191},
  archivePrefix = {arXiv},
  primaryClass = {cs.CV},
}

@misc{li2024llavaov,
  title = {{LL}a{VA}-{O}ne{V}ision: Easy Visual Task Transfer},
  author = {Li, Bo and Zhang, Yuanhan and Guo, Dong and Zhang, Renrui and Li, Feng and Zhang, Hao and Zhang, Kaichen and Li, Yanwei and Liu, Ziwei and Li, Chunyuan},
  year = {2024},
  eprint = {2408.03326},
  archivePrefix = {arXiv},
  primaryClass = {cs.CV},
}

@inproceedings{chen2024internvl,
  title = {{I}ntern{VL}: Scaling up Vision Foundation Models and Aligning for Generic Visual-Linguistic Tasks},
  author = {Chen, Zhe and Wu, Jiannan and Wang, Wenhai and Su, Weijie and Chen, Guo and Xing, Sen and Zhong, Muyan and Zhang, Qinglong and Zhu, Xizhou and Lu, Lewei and Li, Bin and Luo, Ping and Lu, Tong and Qiao, Yu and Dai, Jifeng},
  booktitle = {Proceedings of the {IEEE}/{CVF} Conference on Computer Vision and Pattern Recognition},
  year = {2024},
}

@inproceedings{ye2023ureader,
  title = {{UR}eader: Universal {OCR}-free Visually-situated Language Understanding with Multimodal Large Language Model},
  author = {Ye, Jiabo and Hu, Anwen and Xu, Haiyang and Ye, Qinghao and Yan, Ming and Xu, Yuhao and Qian, Chenliang and Huang, Qi and Zhang, Ji and Wang, Fei and others},
  booktitle = {Findings of the Association for Computational Linguistics: EMNLP 2023},
  year = {2023},
}

@inproceedings{hong2024cogagent,
  title = {Cog{A}gent: A Visual Language Model for {GUI} Agents},
  author = {Hong, Wenyi and Wang, Weihan and Lv, Qingsong and Xu, Jiazheng and Yu, Wenmeng and Ji, Junhui and Wang, Yan and Wang, Zihan and Dong, Yuxiao and Ding, Ming and Tang, Jie},
  booktitle = {Proceedings of the {IEEE}/{CVF} Conference on Computer Vision and Pattern Recognition},
  year = {2024},
}

@inproceedings{mathew2021docvqa,
  title = {{D}oc{VQA}: A Dataset for {VQA} on Document Images},
  author = {Mathew, Minesh and Karatzas, Dimosthenis and Jawahar, C. V.},
  booktitle = {Proceedings of the {IEEE}/{CVF} Winter Conference on Applications of Computer Vision},
  year = {2021},
  eprint = {2007.00398},
  archivePrefix = {arXiv},
}

@inproceedings{mathew2022infovqa,
  title = {Infographic{VQA}},
  author = {Mathew, Minesh and Bagal, Viraj and Tito, Rub{\`e}n and Karatzas, Dimosthenis and Valveny, Ernest and Jawahar, C. V.},
  booktitle = {Proceedings of the {IEEE}/{CVF} Winter Conference on Applications of Computer Vision},
  year = {2022},
  eprint = {2104.12756},
  archivePrefix = {arXiv},
}

@misc{ma2024mmlongbench,
  title = {{MML}ong{B}ench-{D}oc: Benchmarking Long-context Document Understanding with Visualizations},
  author = {Ma, Yubo and Zang, Yuhang and Chen, Liangyu and Chen, Meiqi and Jiao, Yizhu and Li, Xinze and Lu, Xinyuan and Liu, Ziyu and Ma, Yan and Dong, Xiaoyi and Zhang, Pan and Pan, Liangming and Jiang, Yu-Gang and Wang, Jiaqi and Cao, Yixin and Sun, Aixin},
  year = {2024},
  eprint = {2407.01523},
  archivePrefix = {arXiv},
  primaryClass = {cs.CV},
}

@inproceedings{lewis2020rag,
  title = {Retrieval-Augmented Generation for Knowledge-Intensive {NLP} Tasks},
  author = {Lewis, Patrick and Perez, Ethan and Piktus, Aleksandra and Petroni, Fabio and Karpukhin, Vladimir and Goyal, Naman and K{\"u}ttler, Heinrich and Lewis, Mike and Yih, Wen-tau and Rockt{\"a}schel, Tim and Riedel, Sebastian and Kiela, Douwe},
  booktitle = {Advances in Neural Information Processing Systems},
  year = {2020},
}

@inproceedings{khattab2020colbert,
  title = {{C}ol{BERT}: Efficient and Effective Passage Search via Contextualized Late Interaction over {BERT}},
  author = {Khattab, Omar and Zaharia, Matei},
  booktitle = {Proceedings of the 43rd International {ACM} {SIGIR} Conference on Research and Development in Information Retrieval},
  year = {2020},
}

@inproceedings{hu2022lora,
  title = {{L}o{RA}: Low-Rank Adaptation of Large Language Models},
  author = {Hu, Edward J. and Shen, Yelong and Wallis, Phillip and Allen-Zhu, Zeyuan and Li, Yuanzhi and Wang, Shean and Wang, Lu and Chen, Weizhu},
  booktitle = {International Conference on Learning Representations},
  year = {2022},
}

@inproceedings{zhai2023siglip,
  title = {Sigmoid Loss for Language Image Pre-Training},
  author = {Zhai, Xiaohua and Mustafa, Basil and Kolesnikov, Alexander and Beyer, Lucas},
  booktitle = {Proceedings of the {IEEE}/{CVF} International Conference on Computer Vision},
  year = {2023},
}

@misc{beyer2024paligemma,
  title = {{P}ali{G}emma: A versatile 3B {VLM} for transfer},
  author = {Beyer, Lucas and Steiner, Andreas and Pinto, Andr{\'e} Susano and Kolesnikov, Alexander and Wang, Xiao and Salz, Daniel and Neumann, Marcus and Alabdulmohsin, Ibrahim and Tschannen, Michael and Bugliarello, Emanuele and others},
  year = {2024},
  eprint = {2407.07726},
  archivePrefix = {arXiv},
}

@misc{yao2024minicpmv,
  title = {{M}ini{CPM}-{V}: A {GPT}-4{V} Level {MLLM} on Your Phone},
  author = {Yao, Yuan and Yu, Tianyu and Zhang, Ao and Wang, Chongyi and Cui, Junbo and Zhu, Hongji and Cai, Tiannan and Li, Haoyu and Zhao, Weilin and He, Zhihui and others},
  year = {2024},
  eprint = {2408.01800},
  archivePrefix = {arXiv},
}

@inproceedings{kim2022donut,
  title = {{OCR}-free Document Understanding Transformer},
  author = {Kim, Geewook and Hong, Teakgyu and Yim, Moonbin and Nam, JeongYeon and Park, Jinyoung and Yim, Jinyeong and Hwang, Wonseok and Yun, Sangdoo and Han, Dongyoon and Park, Seunghyun},
  booktitle = {European Conference on Computer Vision},
  year = {2022},
}

@inproceedings{lee2023pix2struct,
  title = {Pix2Struct: Screenshot Parsing as Pretraining for Visual Language Understanding},
  author = {Lee, Kenton and Joshi, Mandar and Turc, Iulia and Hu, Hexiang and Liu, Fangyu and Eisenschlos, Julian and Khandelwal, Urvashi and Shaw, Peter and Chang, Ming-Wei and Toutanova, Kristina},
  booktitle = {International Conference on Machine Learning},
  year = {2023},
}

@misc{bai2023qwenvl,
  title = {Qwen-{VL}: A Versatile Vision-Language Model for Understanding, Localization, Text Reading, and Beyond},
  author = {Bai, Jinze and Bai, Shuai and Yang, Shusheng and Wang, Shijie and Tan, Sinan and Wang, Peng and Lin, Junyang and Zhou, Chang and Zhou, Jingren},
  year = {2023},
  eprint = {2308.12966},
  archivePrefix = {arXiv},
}

@misc{liu2023monkey,
  title = {Monkey: Image Resolution and Text Label Are Important Things for Large Multi-modal Models},
  author = {Liu, Zhang and Li, Chunyuan and Wu, Jiannan and Zhang, Qinglong and Wang, Wenhai and Su, Weijie and Chen, Zhe and Dai, Jifeng and Qiao, Yu},
  year = {2024},
  eprint = {2311.06607},
  archivePrefix = {arXiv},
}

@misc{lu2024deepseekvl,
  title = {Deep{S}eek-{VL}: Towards Real-World Vision-Language Understanding},
  author = {Lu, Haoyu and Liu, Wen and Zhang, Bo and Wang, Bingxuan and Dong, Kai and Liu, Bo and Sun, Jingxiang and Ren, Tongzheng and Li, Zhuoshu and Yang, Hao and others},
  year = {2024},
  eprint = {2403.05525},
  archivePrefix = {arXiv},
}

@misc{laurencon2024idefics2,
  title = {What matters when building vision-language models?},
  author = {Lauren{\c{c}}on, Hugo and Tronchon, L{\'e}o and Cord, Matthieu and Sanh, Victor},
  year = {2024},
  eprint = {2405.02246},
  archivePrefix = {arXiv},
}

@misc{jiang2024mixtral,
  title = {Mixtral of Experts},
  author = {Jiang, Albert Q. and Sablayrolles, Alexandre and Roux, Arthur and Mensch, Arthur and Savary, Blanche and Bamford, Chris and Chaplot, Devendra Singh and de las Casas, Diego and Hanna, Emma Bou and Bressand, Florian and others},
  year = {2024},
  eprint = {2401.04088},
  archivePrefix = {arXiv},
}

@misc{openai2024gpt4o,
  title = {{GPT}-4o System Card},
  author = {{OpenAI}},
  year = {2024},
  eprint = {2410.21276},
  archivePrefix = {arXiv},
}

@misc{anthropic2024claude,
  title = {The {C}laude 3 Model Family: Opus, Sonnet, Haiku},
  author = {{Anthropic}},
  year = {2024},
  howpublished = {https://www.anthropic.com/news/claude-3-family},
}

@misc{team2024gemini,
  title = {Gemini 1.5: Unlocking multimodal understanding across millions of tokens of context},
  author = {{Gemini Team}},
  year = {2024},
  eprint = {2403.05530},
  archivePrefix = {arXiv},
}

@inproceedings{sourati2026ladrag,
  title = {{LAD}-{RAG}: Layout-aware Dynamic {RAG} for Visually-Rich Document Understanding},
  author = {Sourati, Zhivar and Wang, Zheng and Liu, Marianne Menglin and Hu, Yazhe and Guo, Mengqing and Bharadwaj, Sujeeth and Han, Kyu J. and Sheng, Tao and Ravi, Sujith and Dehghani, Morteza and Roth, Dan},
  booktitle = {Proceedings of the 64th Annual Meeting of the Association for Computational Linguistics (Volume 1: Long Papers)},
  year = {2026},
  url = {https://aclanthology.org/2026.acl-long.724/},
}

@inproceedings{mara2026,
  title = {{MARA}: A Multimodal Adaptive Retrieval-Augmented Framework for Document Question Answering},
  author = {Wu, Hui and Zhai, Haoquan and Li, Yuchen and Cai, Hengyi and Zhang, Peirong and Zhang, Yidan and Wang, Lei and Wang, Chunle and Hou, Yingyan and Wang, Shuaiqiang and Yin, Dawei},
  booktitle = {Proceedings of the 33rd {ACM} International Conference on Multimedia},
  year = {2025},
  eprint = {2604.16313},
  archivePrefix = {arXiv},
}

@misc{magerag2026,
  title = {{MAGE}-{RAG}: Multigranular Adaptive Graph Evidence for Agentic Multimodal {RAG} in Long-Document {QA}},
  author = {Zuo, Yilong and Li, Xunkai and Yuan, Jing and Dai, Qiangqiang and Qin, Hongchao and Li, Ronghua},
  year = {2026},
  eprint = {2606.15906},
  archivePrefix = {arXiv},
  primaryClass = {cs.CL},
}

@misc{mmr22026,
  title = {Reason Before You Retrieve: Agentic Planning for Multimodal {RAG}},
  author = {Yang, Tianyu and Simon, Shir and Li, Zhenzhen and Cheng, Minhao and Zhang, Xiangliang},
  year = {2026},
  eprint = {2607.22643},
  archivePrefix = {arXiv},
  primaryClass = {cs.CL},
}

@misc{pulsar2026,
  title = {{PULSAR}: Pooled Unified Late-Interaction Search and Retrieval for Enterprise Visual Document {RAG}},
  author = {Constable, Benjamin and Roy, Anup and Sharma, Vishal and Upadhyay, Rishabh Gyanendra and Mills, Robin and Millar, Aidan Philip},
  year = {2026},
  eprint = {2608.28572},
  archivePrefix = {arXiv},
  primaryClass = {cs.IR},
  note = {Accepted at EMNLP 2026 (Industry Track)},
}

@misc{nemotron2026,
  title = {Nemotron {C}ol{E}mbed {V}2: Top-Performing Late Interaction Embedding Models for Visual Document Retrieval},
  author = {de Souza P. Moreira, Gabriel and Ak, Ronay and Xu, Mengyao and Holworthy, Oliver and Schifferer, Benedikt and Yu, Zhiding and Babakhin, Yauhen and Osmulski, Radek and Cai, Jiarui and Chesler, Ryan and Liu, Bo and Oldridge, Even},
  year = {2026},
  eprint = {2602.03992},
  archivePrefix = {arXiv},
  primaryClass = {cs.IR},
}

@misc{anchorfold2026,
  title = {{A}nchor{F}old: A Focus-Then-Fold Framework via Recursive Attention Propagation for Efficient Multi-Vector Visual Document Retrieval},
  author = {Zuo, Haoyu and Yan, Yibo and Zou, Xin and Liu, Shuliang and Cao, Yi and Ou, Mingdong and Hu, Xuming},
  year = {2026},
  eprint = {2608.08732},
  archivePrefix = {arXiv},
  primaryClass = {cs.IR},
}

@misc{colsnap2026,
  title = {{C}ol{SNAP}: Spatial {M}atryoshka Training for Multi-Granularity Visual Document Retrieval},
  author = {Singha Roy, Trishan and Acharya, Arkadeep and Kumar, Vishwajeet and Sen, Jaydeep and Joshi, Sachindra},
  year = {2026},
  eprint = {2608.29951},
  archivePrefix = {arXiv},
  primaryClass = {cs.IR},
}

@inproceedings{karpukhin2020dpr,
  title = {Dense Passage Retrieval for Open-Domain Question Answering},
  author = {Karpukhin, Vladimir and Oguz, Barlas and Min, Sewon and Lewis, Patrick and Wu, Ledell and Edunov, Sergey and Chen, Danqi and Yih, Wen-tau},
  booktitle = {Proceedings of the 2020 Conference on Empirical Methods in Natural Language Processing},
  year = {2020},
}

@misc{xiao2023cpack,
  title = {C-{P}ack: Packaged Resources To Advance General {C}hinese Embedding},
  author = {Xiao, Shitao and Liu, Zheng and Zhang, Peitian and Muennighoff, Niklas},
  year = {2023},
  eprint = {2309.07597},
  archivePrefix = {arXiv},
}

@misc{lee2024nvembed,
  title = {{NV}-{E}mbed: Improved Techniques for Training {LLM}s as Generalist Embedding Models},
  author = {Lee, Chankyu and Lin, Rajarshi and Siemens, Jeremy and others},
  year = {2024},
  eprint = {2405.17428},
  archivePrefix = {arXiv},
}

@inproceedings{radford2021clip,
  title = {Learning Transferable Visual Models From Natural Language Supervision},
  author = {Radford, Alec and Kim, Jong Wook and Hallacy, Chris and Ramesh, Aditya and Goh, Gabriel and Agarwal, Sandhini and Sastry, Girish and Askell, Amanda and Mishkin, Pamela and Clark, Jack and Krueger, Gretchen and Sutskever, Ilya},
  booktitle = {Proceedings of the 38th International Conference on Machine Learning},
  year = {2021},
}

@misc{dong2024internlmxc2,
  title = {{I}ntern{LM}-{X}{C}omposer2: Mastering Free-form Text-{I}mage Composition and Comprehension in Vision-Language Large Model},
  author = {Dong, Xiaoyi and Zhang, Pan and Zang, Yuhang and Cao, Yuhang and Wang, Bin and Ouyang, Linke and Wei, Xilin and Zhang, Songyang and Duan, Haodong and Cao, Maosong and others},
  year = {2024},
  eprint = {2401.16420},
  archivePrefix = {arXiv},
}

@misc{yao2024minicpmv25,
  title = {{M}ini{CPM}-{L}lama3-{V} 2.5: A {GPT}-4{V} Level Multimodal {LLM} on Your Phone},
  author = {Yao, Yuan and Yu, Tianyu and Zhang, Ao and others},
  year = {2024},
  howpublished = {OpenBMB},
}

@inproceedings{liu-etal-2023-matcha,
  title = {{M}at{C}ha: Enhancing Visual Language Pretraining with Math Reasoning and Chart Derendering},
  author = {Liu, Fangyu and Piccinno, Francesco and Krichene, Syrine and Pang, Chenxi and Lee, Kenton and Joshi, Mandar and Altun, Yasemin and Collier, Nigel and Eisenschlos, Julian},
  booktitle = {Proceedings of the 61st Annual Meeting of the Association for Computational Linguistics (Volume 1: Long Papers)},
  year = {2023},
  address = {Toronto, Canada},
  publisher = {Association for Computational Linguistics},
  pages = {12756--12770},
}

@inproceedings{saad-falcon-etal-2024-ares,
  title = {{ARES}: An Automated Evaluation Framework for Retrieval-Augmented Generation Systems},
  author = {Saad-Falcon, Jon and Khattab, Omar and Potts, Christopher and Zaharia, Matei},
  booktitle = {Proceedings of the 2024 Conference of the North American Chapter of the Association for Computational Linguistics: Human Language Technologies (Volume 1: Long Papers)},
  year = {2024},
  address = {Mexico City, Mexico},
  publisher = {Association for Computational Linguistics},
  pages = {338--354},
}

@article{robertson2009bm25,
  title = {The Probabilistic Relevance Framework: {BM}25 and Beyond},
  author = {Robertson, Stephen and Zaragoza, Hugo},
  journal = {Foundations and Trends in Information Retrieval},
  volume = {3},
  number = {4},
  pages = {333--389},
  year = {2009},
}

@misc{chen2024internvl25,
  title = {Expanding Performance Boundaries of Open-Source Multimodal Models with Model, Data, and Test-Time Scaling},
  author = {Chen, Zhe and Wang, Weiyun and Cao, Yue and Liu, Yangzhou and Gao, Zhangwei and Cui, Erfei and Zhu, Jinguo and Ye, Shenglong and Tian, Hao and Liu, Zhaoyang and others},
  year = {2024},
  eprint = {2412.05271},
  archivePrefix = {arXiv},
}
\bibliographystyle{colm2026_conference}

\end{document}